\documentclass[runningheads]{llncs}
\usepackage{graphicx}
\usepackage{amssymb}
\usepackage{multirow}
\usepackage{booktabs}
\usepackage{floatrow}
\newfloatcommand{capbtabbox}{table}[][\FBwidth]

\usepackage{blindtext}

\begin{document}
\title{Enriching Conversation Context in Retrieval-based Chatbots}
%
%
\author{Amir Vakili \and
Azadeh Shakery}
\authorrunning{A. Vakili et al.}
%
\institute{
University of Tehran
\email{\{a\_vakili,shakery\}@ut.ac.ir}\\
}
\maketitle              
\begin{abstract}
Work on retrieval-based chatbots, like most sequence pair matching tasks, can be divided into \textit{Cross-encoders}  that perform word matching over the pair, and \textit{Bi-encoders} that encode the pair separately. The latter has better performance, however since candidate responses cannot be encoded offline, it is also much slower.
Lately, multi-layer transformer architectures pre-trained as language models have been used to great effect on a variety of natural language processing and information retrieval tasks.
Recent work has shown that these language models can be used in text-matching scenarios to create Bi-encoders that perform almost as well as Cross-encoders while having a much faster inference speed.
In this paper, we expand upon this work by developing a sequence matching architecture that
utilizes the entire training set as a makeshift knowledge-base during inference.
We perform detailed experiments demonstrating that this architecture can be used to further improve Bi-encoders performance while still maintaining a relatively high inference speed.

\end{abstract}
\section{Introduction \& Related Works}

Previous literature divides conversational agents into two groups namely task-oriented dialogue systems, that are designed to fulfil a single purpose within a vertical domain such as purchasing an airline ticket, and non-task-oriented chatbots, that are designed to have natural and meaningful conversations with humans on open domain topics \cite{gao2019neural}. Non-task-oriented chatbots can also be trained for information-seeking tasks if they are given the right data. To achieve this, existing methods either employ retrieval based \cite{lowe2015ubuntu,wu2017sequential,dong2018enhance,tao2019multi} or generative \cite{lowe2017training,serban2017hierarchical} conversational agents. Retrieval-based systems select a response from candidates retrieved from chat logs according to how well they match the current conversation context as opposed to generative systems which synthesise new sentences based on the context. As such retrieval-based systems enjoy the advantage of having fluent and informative responses since these were originally written by humans. Retrieval-based chatbots have also been used in real-world products such as Microsoft XiaoIce \cite{shum2018eliza} and Alibaba Group's Alime Assist \cite{feng2017alime}.

Retrieval-based chatbots, like most sequence matchers, are either Bi-Encoders that encode the input pair separately and compare the resulting vectors (also known as representation-focused) or Cross-Encoders that perform matching between all the words in the input pair (also known as interaction-focused) \cite{guo2016deep}. Cross-encoders usually give better results however their word-by-word matching gives rise to higher computational complexity \cite{shen2018deconvolutional}. Moreover, In Bi-encoders candidates response representations are independent of the conversation context so they can be pre-computed offline and cached, thus greatly speeding up the inference process especially if the pool of candidate responses is large \cite{humeau2019real}.

Recently, the introduction of language models pre-trained on large corpora has had a resounding impact on the field of natural language processing \cite{devlin2019bert,liu2019multi} and has been slowly making its way into the conversational information retrieval community \cite{aliannejadi2019asking}. By fine-tuning these models, researchers have achieved state-of-the-art results on various tasks including sequence matching.
These language models usually consist of several transformer layers and can be used in both Bi-Encoder and Cross-Encoder architectures. The Cross-Encoders typically concatenate a sequence pair and feed it to the language model to get a single vector, while Bi-Encoders will calculate a representation for each and then compare the resulting vectors. Recent work showed that Bi-Encoders utilizing BERT \cite{devlin2019bert} can be very effective for response selection and can be further enhanced to get their results closer to Cross-Encoders while still maintaining their fast inference time \cite{humeau2019real}.

In this paper, we expand upon this work by proposing another possible modification to the BERT Bi-Encoders for use in the response retrieval setting which we refer to as \textit{Context Enrichment}. Instead of only comparing the candidate response and conversation context representations, we compare the conversation context to other conversation contexts found in the training set most similar to the candidate response. We expect this change to improve performance as information seeking dialogue systems can cover a vast array of very specific topics and matching a response vector to a particular set of related conversations can help the model better determine whether it is relevant to the conversation context at hand. Essentially we treat the training set as a makeshift knowledge-base. Since conversation representations in the training set and their similarities to candidate responses can be pre-computed and cached offline, this will not add significant overhead to the prediction process.

This approach is similar to \cite{ganhotra2019knowledge} which proposed finding conversations similar to the current conversation and concatenating their corresponding responses to the current context. This technique can works with most methods including our own as it can be considered as a form of data augmentation. Also, our method does not require searching chat logs during inference time as vector similarities between training contexts and candidate responses can be computed and cached offline. Our approach is also inspired by \cite{ai2018learning} who proposed learning global context from a list of retrieved documents in the learning-to-rank setting instead of ranking each document in isolation.

The overall goal of this paper is to show sentence representations made with transformer architectures carry enough information to be used for knowledge enhancement in retrieval-based chatbots. We conduct experiments comparing the new approach to the typical BERT Bi-Encoder and also Cross-Encoder and Bi-Encoder methods that do not use pre-trained transformers.
The experiments are conducted on an existing information-seeking conversational dataset. The experiments show that new architecture improves upon the baseline without greatly affecting its inference speed.

The rest of this paper is organized as follows: In section \ref{sec:method} we explain the task and our proposed method in greater detail. Sections \ref{sec:exp} and \ref{sec:res} are dedicated to detailing experiment settings and obtained results.


\section{Method} \label{sec:method}
In this section, we explain the task in greater detail and provide an overview of our proposed method.

\subsection{Task Definition}
Response retrieval is a setting in which a conversational agent must select the proper response from a group of candidates given the conversation history up to that point. A given model first ranks these candidates according to its prediction as to how well it fits the conversation context, and then these rankings are evaluated using ranking metrics based on gold annotations.
 
 In this paper, we use the Ubuntu corpus dataset which is extracted from the logs of an online chatroom dedicated to the support of the Ubuntu operating system. The dataset features conversations between two people where one is asking for technical help and the other does their best to aid them \cite{lowe2015ubuntu}. We use the version of the dataset prepared by \cite{xu2017incorporating}.

 The task can be formalized as follows: Suppose we have a dataset \(\mathcal{D}=\{c_i,r_i,y_i\}_{i=1}^N\) where \(c_i=\{t_1,\cdots,t_m\}\) represents the conversational context and \(r_i=\{t_1,\cdots,t_n\}\) represents a candidate response and \(y_i\in\{0,1\}\) is a label. \(y_i=1\) means that \(r_i\) is a suitable choice for \(c_i\). The goal of a model should be to learn a function \(g(c,r)\) that predicts the matching degree between any new context \(c\) and a candidate response \(r\).
 
 
\begin{figure}
    \centering
    \includegraphics[width=\textwidth,keepaspectratio]{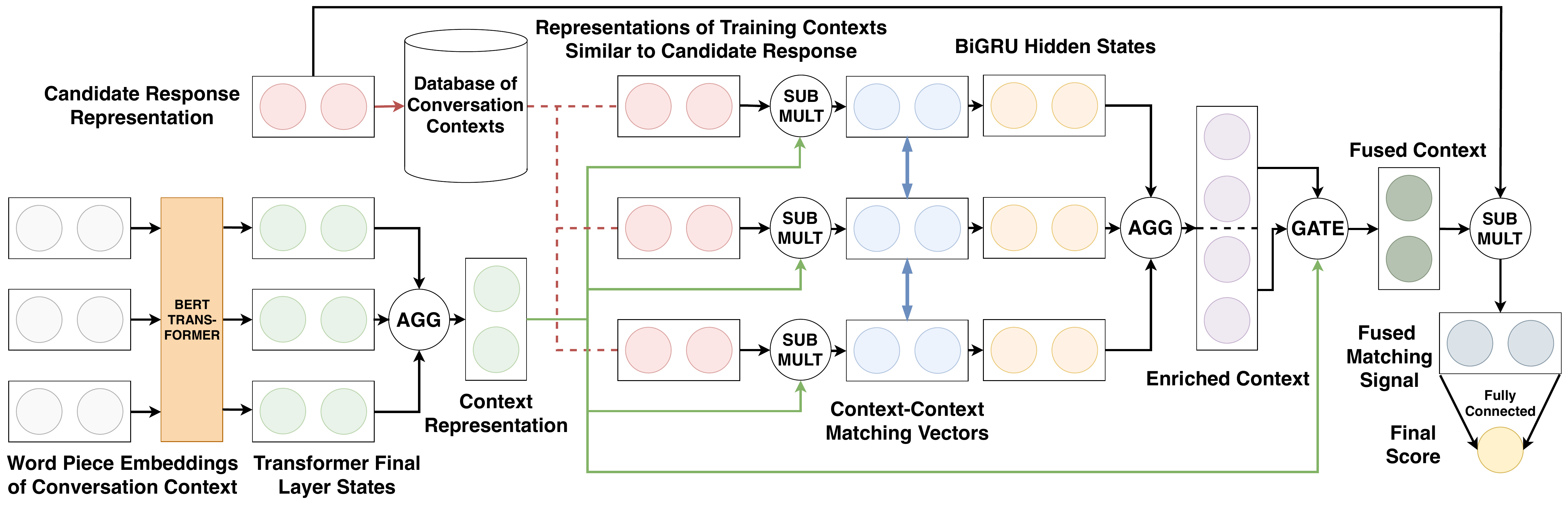}
    \caption{Architecture of proposed method.}
    \label{fig:arch}
\end{figure}{}

\subsection{BERT Bi-Encoder}
The BERT Bi-Encoder \cite{humeau2019real} is the basis of our architecture. It uses BERT, a multi-layer transformer pre-trained as a language model \cite{devlin2019bert}, to compute sentence representations from the conversation context and candidate response separately and then matches these representations together to predict their matching degree.
The two representations are computed as follows:
\[
\begin{array}{cccccc}
     v_c = & reduce(T(context)) &&&
     v_r = & reduce(T(response))
\end{array}
\]
where \(T\) is the BERT transformer, \(reduce\) is a function that aggregates the final hidden states of the transformer into a single vector. Following previous work \(reduce\) selects the first hidden vector (corresponding to the special token [CLS]). 

To compare the two vectors we use a slightly modified version of the \textit{SubMult} function \cite{wang2017compagg} coupled with a two-layer feed-forward network as we found it gave better results than simply using the dot product of the two vectors:
\[
\begin{array}{c}
    SubMult(v_c, v_r) = [v_c, v_r, v_c - v_r, v_c - v_r, (v_c - v_r) * (v_r - v_c)] \\
     \hat{y} = m(v_c, v_r) = W_2(ReLU(W_1(SubMult(v_c, v_r)) \\
\end{array}{}
\]
where \(\hat{y}\) is the predicted matching score for the context response pair; $[\cdot,\cdot]$ is the concatenation operation. The network is trained to minimize cross-entropy loss where the logits are \(m(c,r_0),\cdots,m(c,r_1)\) and \(r_0\) is the correct response. The rest of the candidate responses are selected from other samples in the training batch as this greatly speeds up training \cite{humeau2019real,mazare2018millions}.

\subsection{BERT Bi-Encoder+CE}
Bi-Encoder+CE is our proposed architecture. The BERT Bi-Encoder+CE not only compares a context vector to the candidate responses vector, but it also compares it to $k$ context vectors similar to the response candidate that are retrieved from the training set using cosine similarity. The architecture is depicted in Figure \ref{fig:arch}.

The model first constructs the representation vector $C$ of the conversation context, similar to the regular BERT Bi-Encoder. It then uses the pre-computed candidate response vector $R$ to retrieve a set of $k$ similar contexts $\{C_1^R,\cdots,C_k^R\}$ from the training set, which have also been pre-computed, using cosine similarity. It then compares each of them to the context vector using the SubMult function. The resulting vectors are passed to a bidirectional
GRU \cite{chung2014empirical}.
$$
\begin{array}{ccccccc}
     \hat{C}_i^R =  SubMult(C, C_i^R)  &&,&&
     H_i =  BiGRU( \hat{C}_i^R) \\
\end{array}{}
$$

We use additive attention in order to aggregate the GRU hidden states $\{H_1,\cdots,H_k\}\subset\mathbb{R}^{2d}$  into a single vector $\hat{C} \in \mathbb{R}^{2d}$ as follows: 

$$
\begin{array}{ccccccc}
     a_i =  softmax(W_{13}(tanh(W_{11}H_i + W_{12}R)) &&,&&
     \hat{C} =  \sum_{i=1}^ka_iH_i \\
\end{array}{}
$$

This is so the model can use the most relevant helper contexts to build the enriched context vector.

For the model to learn when to use the enriched context $C_e$ and when to use the regular context, we use the first half of $\hat{C}$ as a control for a gating mechanism. The resulting fused context vector $C_f$ is then compared to the response representation $R$:

$$
\begin{array}{ccc}
     C_e & = & \{\hat{c}_i,\cdots,\hat{c}_{k/2}\},G = \{\hat{c}_{k/2+1},\cdots,\hat{c}_n\} \\
     C_f & = & sigmoid(G)*C_e + (1-sigmoid(G))*C \\
     \hat{y} & = & W_{22}(ReLU(W_{21}SubMult(C_f,R))) \\
\end{array}{}
$$
\section{Experiment Settings} \label{sec:exp}
Our models are implemented in the PyTorch framework \cite{paszke2017pytorch}. For our BERT component, we used the Distilbert implementation available in Huggingface's transformer library\footnote{\url{https://github.com/huggingface/transformers}} since it provides reasonable results despite having only 6 layers of transformers instead of the 12 in the original implementation. We train the networks three times using different seeds and for each model select the one that gave the best results on the development set. We also implemented Dual-LSTM \cite{lowe2015ubuntu} and ESIM \cite{dong2018enhance} as examples of non-transformer-based Bi-Encoder and Cross-Encoders. Our code will be released as open-source. 

To shorten training time we train a BERT Bi-Encoder until convergence using the AdamW optimizer \cite{loshchilov2019adamw} and a learning rate of $5\times10^{-5}$. We fine-tune other networks using it as a starting point with the Adam optimizer\cite{kingma2014adam} and a learning rate of $10^{-4}$. Batch size is 32 for BERT models and 128 for the rest.

% ...

\section{Results and Discussion} \label{sec:res}
The results depicted in Table \ref{table:results} show that the inclusion of training contexts similar to the candidate responses can be beneficial to the response retrieval process. 
It can also be noted that Context Enrichment (CE) adds a relatively small amount of overhead and is still faster than even simple Cross-Encoder architectures.
Table 2 shows that increasing the number of considered contexts similar to the candidate response improves performance but only up to $k=20$. 

Due to limited compute resources we were not able to match the state-of-the-art Bi-Encoder result \cite{humeau2019real}. Their models were trained on 8 16GB GPUs while the results here were obtained using a single 11GB GPU. This means we had to make compromises on the sequence lengths, the number of negative samples and model size. This is especially apparent in our BERT Cross-Encoder result as increasing negative samples to match the Bi-Encoder is infeasible on one GPU.
\begin{table} 
\caption{Comparison of model performance. Starred results are reported from their respective papers. All other models were re-implemented in Pytorch. Our main baseline is the BERT Bi-Encoder, BERT Bi-Encoder+CE metrics that are statistically significant relative to it are marked in bold. We use paired two-tailed t-tests with a p-value$<$0.05. Inference time is average milliseconds to process a single sample on a GPU.
}
    \centering
    \scriptsize
    \setlength\tabcolsep{1.5pt}
    \begin{tabular}{lcccc@{\hskip .1in}cccc}
        \toprule
        \cmidrule(lr){2-6}
                 &  $R_{10}@1$  & $R_{10}@2$  & $R_{10}@5$   & $MRR$   & Inference time\\
        \midrule
        SMN*\cite{wu2017sequential}     & 72.6 & 84.7 & 96.1 & ---  & --- \\
        ESIM*\cite{dong2018enhance}     & 75.9 & 87.2 & 97.3 & 84.8 & --- \\
        MRFN*\cite{tao2019multi}        & 78.6 & 88.6 & 97.6 & ---  & --- \\
        \midrule
        Dual-LSTM\cite{lowe2015ubuntu}  & 54.8 & 73.1 & 91.3 & 70.3   & ---\\
        ESIM \cite{dong2018enhance}     & 74.4 & 85.2 & 96.1 & 83.4  & 9\\
        \midrule
        BERT Bi-Encoder                      & 78.0 & 88.5 & 97.4 & 86.2 &  3\\
        BERT Bi-Encoder+CE                   & \textbf{79.3} & \textbf{89.3} & 97.5 & \textbf{87.0} & 4.5\\
        \midrule
        BERT Cross-Encoder                   & 76.5 & 86.4 & 96.4 & 84.8 & 30 \\
        \bottomrule
    \end{tabular}
    \label{table:results}
\end{table}

\subsection{Ablation Study}

Table 3 depicts how well the model performs with various components altered. These results show that only when we combine the enriched context and the regular context are we able to see significant gains. This means the enriched context vector contains information complementary to the regular context vector. The results also demonstrate the effectiveness of the SubMult function. 

\section{Conclusion and Future Work} \label{sec:conc}
In this paper, we introduced a new architecture for use in retrieval-based chatbots. When predicting the matching degree between a candidate response and a conversation context, the new architecture takes into account contexts within the training set that are possible matches for the candidate response and compares them to the current context. The model improves upon the BERT Bi-Encoder baseline without greatly affecting inference speed. We provide an overview of the performance/speed trade-off between the mentioned architectures. 

The advent of powerful sequence-encoders using transformer architectures gives researchers a new avenue to explore as Bi-Encoders have been somewhat ignored in favour of the more computationally expensive Cross-Encoders. One possible new  approach we intend to explore is to use graph convolution networks \cite{kipf2017graph} to construct a makeshift knowledge-base that can provide a much richer and more structurally sound model for the relations between contexts and responses, similar to work done previously on text classification \cite{yao2019graph}.

\begin{figure}
\scriptsize
\begin{floatrow}
\capbtabbox{%
  \begin{tabular}{lcc}
        \toprule
            $k$ &  Dev $R_{10}@1$  & Test $R_{10}@1$   \\
        \midrule
        0 & 78.1 & 78.0 \\
        2 & 78.7 & 78.5 \\
        5 & 78.6 & 78.6 \\
        10 & 78.9 & 79.0 \\
        15 & 79.0 & 79.1 \\
        \textbf{20} & \textbf{79.2} & \textbf{79.3} \\
        25 & 79.1 & 79.2 \\
        \bottomrule
    \end{tabular}
    \label{table:kparam}
}{%
  \caption{Analysis of hyper-parameter $k$}%
}

\capbtabbox{%
  \begin{tabular}{lcc}
        \toprule
            &  Dev $R_{10}@1$  & Test $R_{10}@1$   \\
        \midrule
        \textbf{Bi-Encoder+CE}      & \textbf{79.2}  & \textbf{79.3}  \\
        -Attention      & 78.9  & 78.8  \\
        -Gate           & 78.0  & 77.9 \\
        \midrule
        Bi-Encoder      & 78.1 & 78.0 \\
        -SubMult        & 76.8 & 76.5 \\
        \bottomrule
    \end{tabular}
    \label{table:ablation}
}{%
  \caption{Ablated model metrics}%
}
\end{floatrow}
\end{figure}

%
%
%
\bibliographystyle{splncs04}
\bibliography{samplepaper}
\end{document}